# Long-Range Feature Propagating for Natural Image Matting

Qinglin Liu
Harbin Institute of Technology
qinglin.liu@outlook.com

Haozhe Xie
Harbin Institute of Technology
hzxie@hit.edu.cn

Shengping Zhang*
Harbin Institute of Technology
s.zhang@hit.edu.cn

Bineng Zhong
Guangxi Normal University
bnzhong@hqu.edu.cn

Rongrong Ji
Xiamen University
rrji@xmu.edu.cn

## ABSTRACT

Natural image matting estimates the alpha values of unknown regions in the trimap. Recently, deep learning based methods propagate the alpha values from the known regions to unknown regions according to the similarity between them. However, we find that more than 50% pixels in the unknown regions cannot be correlated to pixels in known regions due to the limitation of small effective reception fields of common convolutional neural networks, which leads to inaccurate estimation when the pixels in the unknown regions cannot be inferred only with pixels in the reception fields. To solve this problem, we propose Long-Range Feature Propagating Network (LFPNet), which learns the long-range context features outside the reception fields for alpha matte estimation. Specifically, we first design the propagating module which extracts the context features from the downsampled image. Then, we present Center-Surround Pyramid Pooling (CSPP) that explicitly propagates the context features from the surrounding context image patch to the inner center image patch. Finally, we use the matting module which takes the image, trimap and context features to estimate the alpha matte. Experimental results demonstrate that the proposed method performs favorably against the state-of-the-art methods on the AlphaMatting and Adobe Image Matting datasets.

## 1 INTRODUCTION

Natural image matting aims to estimate the alpha mattes (opacity) of the foreground from natural images, which has many potential applications, such as image composition [6, 7], film post-production [16] and live streaming [15]. Mathematically, the observed image $\mathcal{I}$ is modeled as a convex combination of the foreground $\mathcal{F}$ and background $\mathcal{B}$ as

$$\mathcal{I}_i = \alpha_i \mathcal{F}_i + (1 - \alpha_i) \mathcal{B}_i \tag{1}$$

where $\alpha_i$ is the opacity of the foreground at pixel $i$. Since the foreground $\mathcal{F}$, background $\mathcal{B}$, and alpha matte $\alpha$ are unknown, only the observed image $\mathcal{I}$ is known, image matting is an ill-defined problem. To address this problem, most existing methods [3, 5, 8, 12, 13, 15, 19, 25, 26, 28, 41, 44, 45] need a trimap to indicate the known foreground and background regions and unknown regions in the image, which can be roughly categorized into sampling-based methods, propagation-based methods, and learning-based methods.

Sampling-based methods [2, 9, 12, 15, 31, 36, 41] select the best foreground and background color pairs in the known regions for the each pixel in unknown regions to estimate the alpha values based on assumptions on color association. Propagation-based methods [5, 17, 19, 23, 25, 39] propagate the alpha values from the known regions to unknown regions according to the similarity between them. However, all the above methods are sensitive to the foreground and background color distributions overlap that are common in natural images, which leads to unexpected artifacts in the predicted alpha mattes [44]. Recently, learning-based methods [3, 8, 13, 26, 28, 44, 45] use neural networks to learn both the color information and natural structure from well-annotated datasets, which achieve better performance than traditional methods. However, these methods only learn local image features due to the small effective reception fields of neural networks, which ignore the long-range image features outside the reception fields. By analyzing the widely used Adobe Image Matting dataset [44], we find that more than 50% pixels in the unknown regions cannot be correlated to pixels in known regions in the range of the effective reception fields [29], which leads to inaccurate estimation.

To improve existing learning-based image matting methods, we propose the Long-Range Feature Propagating Network (LFPNet) for natural image matting, which learns the long-range features outside the reception fields to help distinguish the foreground and background locally. In particular, it consists of the propagating module and the matting module, as shown in Figure 1. Specifically, given the downsampled context image patch as input, the propagating module first extracts the rich context features around the inner center while avoiding the high computational cost. Then, we devise a center-surround pyramid pooling (CSPP) as the bottleneck of the propagating module, which adopts a new designed center-surrounding pooling and an Atrous Spatial Pyramid Pooling (ASPP) [4] to explicitly propagate the long-range context features from the context image patch to the inner center image patch and perform feature map smoothing. Finally, the matting module takes the input image, trimap and context features from the propagating module to estimate the alpha matte, foreground, and background simultaneously. Benefited from the long-range feature propagating, the proposed LFPNet performs well when the difference between the foreground and the background is small in a local region.

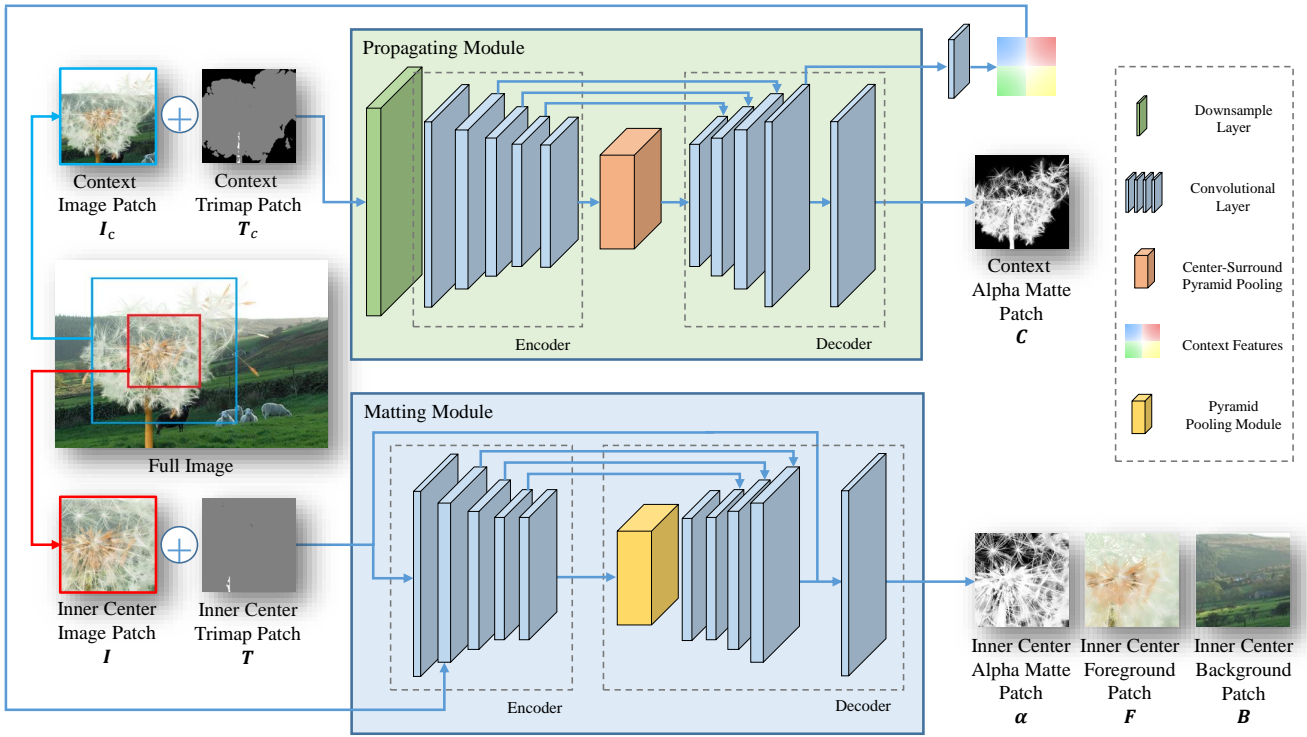

Figure 1: Overview of the proposed Long-Range Feature Propagating Network (LFPNet). LFPNet performs image matting in a patch-based crop-and-stitch manner. The propagating module takes the context image patch $I_c$ and context trimap patch $T_c$ centered on the inner center image patch $I$ and trimap patch $T$ as inputs to predict the context alpha matte $C$ and outputs context features. The matting module takes the inner center image patch $I$, trimap patch $T$, and context features as inputs to predict the inner center alpha matte patch $\alpha$, foreground patch $F$, and background patch $B$.

In summary, the contributions of this paper are threefold:

- We propose Long-Range Feature Propagating Network (LFPNet) for image matting, which learns the long-range features outside the reception fields. LFPNet benefits the alpha matte estimation for the pixels in unknown regions by leveraging more pixels in the known foreground and background regions.
- We present Center-Surround Pyramid Pooling (CSPP) that generates the context features by explicitly propagating the features outside the reception fields to the inner center image patch.
- Experiment results on the AlphaMatting and Adobe Image Matting datasets demonstrate that the proposed method outperforms state-of-the-art methods.

## 2 RELATED WORK

**Sampling-based methods.** Sampling-based methods sample the pixels in known regions near the unknown regions to estimate the possible foreground and background colors for a given pixel. These methods design metrics to use the similarity between the unknown pixels and the known pixels to estimate the alpha value. Mishima [31] proposes the first image matting method based on the foreground and background color sampling. Berman et al. [2] propose to use the known pixels around the unknown pixel to obtain the alpha value of the unknown pixel with a linear function. Ruzon and Tomasi [36] use statistical methods to cluster the foreground and background colors to estimate the foreground and background colors for pixels in the unknown regions. Bayesian Matting [9] estimates the alpha value by using the Bayes' theorem to maximize the posterior probability. Robust Matting [41] considers the spatial information and samples the foreground and background regions that are spatially close to estimate the alpha matte. Shared Matting [15] uses only the pixels in the near boundaries between the known and unknown regions to speed the image matting. Global Matting [18] samples all the pixels in the boundaries to prevent missing important information, but it results in slower speed. Shahrian *et al.* [38] propose to use the distance from the known area to the unknown area to cluster the image into superpixels and then perform sampling. Feng *et al.* [12] avoid the shortcomings of spatial assumptions by sampling the candidate foreground and background colors based on both color and spatial statistics, then use sparse coding to establish an objective function to estimate the alpha value.

**Propagation-based methods.** Propagation-based methods use a priori of local smoothing to formulate a cost function to propagate the alpha value from known regions to unknown regions. Poisson Matting [39] uses Poisson equation to estimate alpha value. Close-form matting [25] first establishes a Color-Line model, based on which a closed-form solution for alpha matte estimation is derived. Spectral matting [23] introduces spectral clustering for the Close-form matting. He *et al.* [19] accelerate the Close-form matting algorithm by increasing the window size of the Laplace matrix to reduce iterations. KNN matting [5] collects K nearest neighbors globally in high-dimensional feature space to solve the image problem and gives a closed-form solution, which increases the speed while maintaining the accuracy of the matting. Information-flow Matting [1] combines both local and non-local affinities to estimate the alpha matte.

**Learning-based methods.** Learning-based methods learn knowledge from the image and annotation to predict the alpha matte. Benefiting from the development of deep learning and image matting datasets, many learning-based methods have emerged. Cho *et al.* [8] propose a deep learning method to refine the alpha mattes from Close-Form matting [25] and KNN matting [5]. DIM [44] provides the first large-scale image matting dataset and introduces the first end-to-end matting module with a refinement module. AlphaGan [30] uses the adversarial loss to improve the accuracy of matting. SampleNet [40] uses the foreground and background information to supervise the network to improve prediction accuracy. AdaMatting [3] refines the trimap while predicting the alpha matte. IndexNet [28] reserves the pooling indices for unpooling operation and improves the gradient accuracy and visual perception of the results. GCAMatting [26] designs a Guided Contextual Attention module to capture contextual affinity information to estimate the alpha matte of semi-transparent objects. FBAMatting [13] designs a network to estimate the foreground color, background color and alpha matte at the same time, and uses a first-order approximation to the Bayesian formula to refine the prediction results. HDMatt [45] introduces a Cross-Patch Contextual module to improve the performance under patch-based inference.

## 3 MOTIVATION

Recently, deep learning based methods propagate the alpha values from the known regions to unknown regions according to the similarity of the two regions. According to the statistics on the Adobe Image Matting dataset [44], we find that more than 50% pixels in the unknown regions cannot be correlated to pixels in known regions due to the limitation of the small effective reception fields, which leads to inaccurate estimation. In this section, we analyze the above observations in detail.

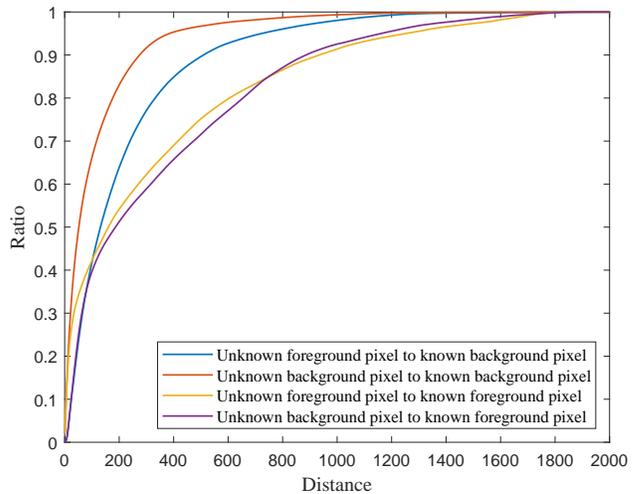

**Figure 2: Ratio of the shortest distance from the pixels in the unknown regions to the foreground or background pixels in the known regions on the Adobe Image Matting dataset [44].**

First, we collect the images whose trimaps contain foreground and background pixels in known regions in the Adobe Image Matting dataset. Then, we classify the pixels in unknown regions into the foreground and background pixels according to their alpha values. Finally, we calculate the Euclidean distance from each pixel in unknown regions to the nearest pixels in the known regions. As shown in Figure 2, the distance between the pixels in unknown regions and the pixels in known regions varies greatly. The distance between the background pixels in the unknown regions and the background pixels in the known regions is the shortest, but half of the pixels are more than 58 pixels away from the nearest known background pixels. Compared to the distance between the background pixels in the unknown regions and the background pixels in the known regions, the distance between the foreground or background pixels in the unknown regions and the foreground pixels in the known regions is farther, and half of the pixels are more than 167 pixels away from the nearest known foreground pixels, which exceeds the effective receptive field [29] size of 75 pixels of the commonly used ResNet-50 [21]. Besides, 25% of the pixels in the unknown regions are more than 500 pixels away from the nearest foreground pixels in the known regions.

Based on the observations, it is necessary to consider the long-range features outside the reception fields in natural image matting.

## 4 THE PROPOSED METHOD

LFPNet follows a patch-based crop-and-stitch manner, which firstly crops an input image and trimap into patches and then estimates the alpha values of each patch. We denote the cropped image patch and trimap patch as inner center image patch $I$ and inner center trimap patch $T$, respectively. To leverage the long-range features, we generate the context image patch $I_c$ and the context trimap patch $T_c$ by enlarging the crop patch $I$ and $T$ by a scale of 2, the context patch has four times the area and context information of the inner center patch.

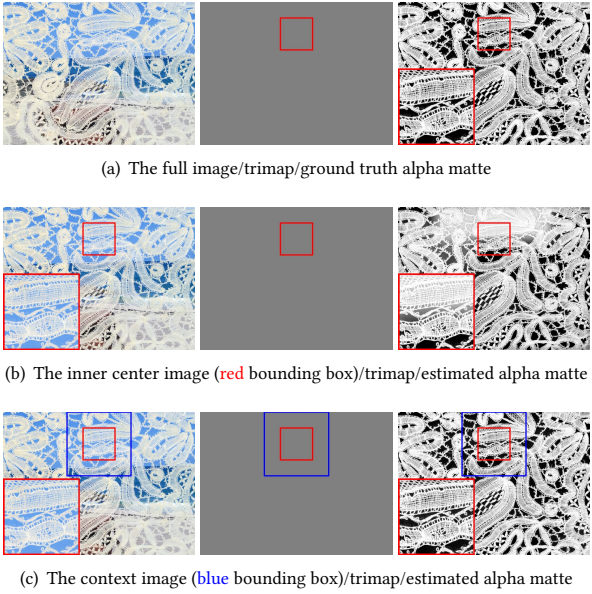

(a) The full image/trimap/ground truth alpha matte

(b) The inner center image (red bounding box)/trimap/estimated alpha matte

(c) The context image (blue bounding box)/trimap/estimated alpha matte

**Figure 3: The illustration of the image, trimap and alpha matte in the Adobe Image Matting dataset. (a) shows the full image, trimap and alpha matte ground truth. (b) shows the inner center image patch, trimap and the predicted alpha matte. (c) shows the context image patch, trimap and the predicted alpha matte.**

The proposed LFPNet consists of two modules, including the propagating module and the matting module. Given the context image patch and the context trimap patch as input, the propagating module generates the context features and the context alpha matte patch $C$. The matting module generates the inner center alpha matte $\alpha$, inner center foreground patch $F$, and inner center background patch $B$ for the inner center image patch $I$ from the context features, the inner center image patch $I$ and the corresponding trimap $T$.

## 4.1 Propagating Module

The propagating module extracts the long-range image features from the context image for alpha matte estimation. The propagating module consists of three components: the context encoder, Center-Surround Pyramid Pooling, the context decoder.

*4.1.1 Context Encoder.* Different from the existing deep learning based methods that take only the inner center image patch $I$ as input, the context encoder takes the context image patch $I_c$ as input. Figures 3(b) and 3(c) give the comparison of the estimated alpha mattes with only inner center image and the context image, which indicates that the context features are necessary for the alpha matte estimation. Intuitively, the inner center image patch only contains a small part of the image, which makes it difficult for the networks to distinguish the foreground and background of the image.

The context encoder aims to extract context features with a small computational cost, which enables the network to leverage the context features for the alpha matte estimation. To reduce the computational cost of the context encoder, the input context image patches are downsampled by bicubic interpolation and convolution operations. Next, the ResNet-50 [21] is adopted as the backbone to extract the context features from the context image patch. As shown in Figure 3, a larger reception field contains more information, which helps to distinguish the foreground and background. To further enlarge the effective reception field [29], we replace the convolutional layers of *block*3 and *block*4 in ResNet-50 with the dilated convolutional layers whose dilation rates are 2 and 4, respectively. Consequently, the context encoder generates the context feature map for the context image patch.

*4.1.2 Center-Surround Pyramid Pooling.* To incorporate the features from the context image patch to the inner center image patch, we propose Center-Surround Pyramid Pooling (CSPP), which explicitly performs long-range feature propagating at multiple feature scales. The proposed CSPP contains two components: Center-Surround Pooling and Atrous Spatial Pyramid Pooling [4], as illustrated in Figure 4.

**Center-Surround Pooling (CSP)** aims to propagate the context feature map from the known regions to the near unknown regions. Specifically, CSP first divides the context feature map into $1^2$, $2^2$, $3^2$, and $6^2$ blocks, which establishes the connections between the features of the unknown regions and known regions at different distances. Next, the multi-scale block-wise average pooling is applied on each block to obtain the average features of the block. Since the area of the surrounding regions is significantly larger than the area of the inner center image patch, the average features of most blocks are dominated by the features of the surrounding regions. Therefore, the features are propagated from the context image patch to the inner center image patch. Finally, the average features in each block are projected by $1 \times 1$ convolutional layers, which are then concatenated with the context feature map to generate the coarsely propagated features.

**Atrous Spatial Pyramid Pooling (ASPP)** is adopted to smooth the coarsely propagated features while further enlarging the reception fields. Specifically, ASPP consists of a $1 \times 1$ convolutional layer, a global average pooling layer followed by a $1 \times 1$ convolutional layer, and four $3 \times 3$ dilated convolutional layers with dilation rates of 3, 7, 12, 18. The feature maps generated by the six layers are concatenated and fed into the context decoder.

*4.1.3 Context Decoder.* The context decoder aims to upsample the context features to the original size while recovering the low-level features lost during the downsampling process of the context encoder. Specifically, the context decoder consists of four convolutional layers followed by up-sampling layers and generates the context features for the context image patch. At the end of the context decoder, an additional convolutional layer is introduced to estimate the alpha matte $C$ of the context image patch.

## 4.2 Matting Module

To estimate the alpha matte for the input image associated with the trimap, we present the matting module following DIM [44]. The matting module adopts a U-Net style architecture, which contains the matting encoder for feature extraction and the matting decoder for alpha matte, foreground, and background estimation.

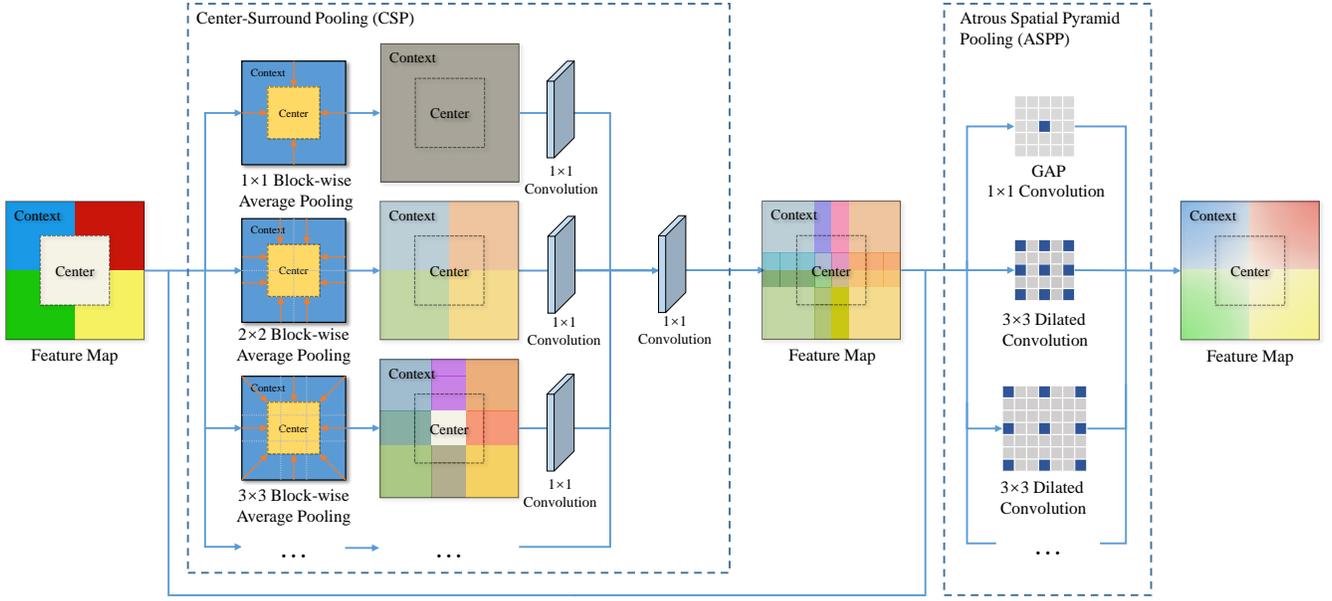

Figure 4: The architecture of Center-Surround Pyramid Pooling (CSPP). The center-surround pooling (CSP) uses the multi-scale block-wise average pooling to generate the coarsely propagated features. Atrous Spatial Pyramid Pooling (ASPP) uses multiple dilated convolutional layers to further smooth the coarsely propagated features.

*4.2.1 Matting Encoder.* The matting encoder aims to extract multi-scale image features from the inner center image patch and context features. To this aim, we also adopt the ResNet-50 [21] as the backbone network due to its powerful feature extraction capability. To preserve the low-level image features of the image, we replace the shallow stem with a deep stem composed of multiple $3 \times 3$ convolutional layers following Res2Net [14]. To preserve spatial information of feature maps, we replace the convolutional layers of *block*3 and *block*4 with dilated convolutional layers with dilation rates of 2 and 4, respectively.

*4.2.2 Matting Decoder.* The matting decoder utilizes the multi-scale image features extracted by the matting encoder to estimate alpha matte. We adopt the residual blocks with bilinear upsampling layers and skip-connections as the basic building blocks, Specifically, we first adopt Pyramid Pooling Module [46] to learn to fuse semantic information on multiple scales, which helps to make full use of features of different resolution image patches. Then, we stack four residual blocks with bilinear upsampling layers to upsample the feature maps to the resolution of the input image patch. To refines the details of the upsampled feature maps, we use skip-connections to incorporate the low-level features. Finally, we stack three convolutional layers to estimate the foreground $F$, background $B$ the alpha matte $\alpha$ following FBAMatting [13].

## 4.3 Loss Function

We use the propagating loss and matting loss to train the propagating module and the matting module, respectively.

**Propagating Loss.** Given the estimated context alpha matte $C$ and the corresponding ground truth $C^{gt}$, the propagating loss $\mathcal{L}_p$ is defined as

$$\mathcal{L}_p = \frac{1}{|\mathcal{T}_c^U|} \sum_{i \in \mathcal{T}_c^U} |C_i - C_i^{gt}| \quad (2)$$

where $\mathcal{T}_c$ is the set of unknown pixels in the context trimap patch $T_c$.

**Matting Loss.** Given the estimated inner center alpha matte patch $\alpha$, inner center foreground patch $F$, the inner center background patch $B$, and the corresponding ground truth $\alpha^{gt}$, $F^{gt}$, $B^{gt}$, the matting loss can be formulated as

$$\mathcal{L}_m = \lambda_\alpha \mathcal{L}_\alpha + \lambda_{FB} \mathcal{L}_{FB} \quad (3)$$

where $\mathcal{L}_\alpha$ and $\mathcal{L}_{FB}$ are the alpha matte loss and the foreground-background color loss, respectively. $\lambda_\alpha$ and $\lambda_{FB}$ are the weights for the two terms.

The alpha matte loss $\mathcal{L}_\alpha$ is computed as

$$\mathcal{L}_\alpha = \mathcal{L}_w + \mathcal{L}_I + \mathcal{L}_{L\alpha} \quad (4)$$

where the weight alpha loss $\mathcal{L}_w$, composite loss $\mathcal{L}_I$, and Laplacian alpha loss $\mathcal{L}_{L\alpha}$ are defined as

$$\mathcal{L}_w = \frac{max(1, \sqrt{\frac{\sum_{i \in \mathcal{T}^U}}{\gamma}})}{|\mathcal{T}^U|} \sum_{i \in \mathcal{T}^U} |\alpha_i - \alpha_i^{gt}| \quad (5)$$

$$\mathcal{L}_I = |\alpha F^{gt} + (1-\alpha)B^{gt} - I| \quad (6)$$

$$\mathcal{L}_{L\alpha} = \text{Lap}(\alpha, \alpha^{gt}) \quad (7)$$

where $\mathcal{T}^U$ is a set of unknown pixels in the inner center trimap $T$, $\gamma$ is the coefficient used to increase the weight for trimaps containing large unknown regions. Lap represents the Laplacian pyramid cost, which is defined as

$$\text{Lap}(x, y) = \sum_j 2^j |L^j(x), L^j(y)| \quad (8)$$

where $L^j(x)$ and $L^j(y)$ are the $j$-th level of the Laplacian pyramid representations of $x$ and $y$, respectively.

Table 1: The average ranking results on the AlphaMatting dataset. Note that S, L, U stand for small trimap, large trimap, and user trimap, respectively. The best results are highlighted in bold.

| Methods | SAD | | | | MSE | | | | Grad | | | |
| --- | --- | --- | --- | --- | --- | --- | --- | --- | --- | --- | --- | --- |
| | Overall | S | L | U | Overall | S | L | U | Overall | S | L | U |
| LFPNet (Ours) | **3.6** | **3.1** | **2.8** | **4.9** | **3.2** | **3.3** | **2.1** | **4.3** | **2.3** | **2.3** | **2.1** | **2.5** |
| HDMatt [45] | 9.5 | 11.6 | 8.1 | 8.8 | 9.6 | 12.4 | 8.1 | 8.4 | 7.8 | 8.9 | 6.1 | 8.5 |
| Background Matting [37] | 11.8 | 9.3 | 10 | 16.3 | 11.1 | 7.8 | 9.3 | 16.4 | 10.7 | 6.9 | 9 | 16.1 |
| AdaMatting [3] | 11.9 | 10.6 | 10.9 | 14.3 | 12.6 | 10.1 | 11.5 | 16.1 | 12.1 | 8.3 | 9.9 | 18.3 |
| SampleNet Matting [40] | 12.3 | 9.9 | 12.1 | 14.8 | 13 | 9.3 | 13.5 | 16.3 | 13.9 | 9.6 | 11.6 | 20.5 |
| GCA Matting [26] | 13.5 | 14.8 | 10.8 | 15.1 | 14.4 | 14.5 | 13 | 15.6 | 12.5 | 12.8 | 11 | 13.8 |
| Deep Matting [44] | 15.2 | 16.6 | 14.3 | 14.8 | 18.5 | 17.3 | 17.3 | 20.9 | 23 | 20.1 | 19.5 | 29.4 |
| IndexNet Matting [28] | 18.8 | 21.4 | 17.6 | 17.5 | 22.5 | 25.1 | 21 | 21.3 | 17.7 | 16.6 | 16.4 | 20.1 |

The foreground-background color loss $\mathcal{L}_{FB}$ consists of the foreground background reconstruction loss $\mathcal{L}_{FBR}$, foreground background composite loss $\mathcal{L}_{FBC}$, and the Laplacian foreground background loss $\mathcal{L}_{LFB}$, which can be computed as

$$\mathcal{L}_{FB} = \mathcal{L}_{FBR} + \mathcal{L}_{FBC} + \mathcal{L}_{LFB} \quad (9)$$

where

$$\mathcal{L}_{FBR} = \frac{1}{|\mathcal{T}^{FU}|} \sum_{i \in \mathcal{T}^{FU}} |F_i - F_i^{gt}| + \frac{1}{|\mathcal{T}^{BU}|} \sum_{i \in \mathcal{T}^{BU}} |B_i - B_i^{gt}| \quad (10)$$

$$\mathcal{L}_{FBC} = |\alpha^{gt} F + (1 - \alpha^{gt}) B - I| \quad (11)$$

$$\mathcal{L}_{LFB} = \text{Lap}(F, F^{gt}) + \text{Lap}(B, B^{gt}) \quad (12)$$

where $\mathcal{T}^{FU}$ and $\mathcal{T}^{BU}$ are the set of foreground and unknown pixels and the set of background and unknown pixels in the inner center trimap $T$, respectively.

## 5 EXPERIMENTS

### 5.1 Datasets

**AlphaMatting.** AlphaMatting [35] is an image matting test dataset that consists of 8 real-world testing images for the online benchmark. Each testing image is with three different trimaps (i.e. "small", "large", "user") for evaluation.

**Adobe Image Matting.** Adobe Image Matting [44] is an image matting test dataset that consists of 1,000 high-resolution testing images. These images are synthesized from 50 foreground images and 1,000 background images from the PASCAL VOC dataset [11]. Each testing image has a unique trimap.

### 5.2 Implementation Details

The proposed LFPNet is implemented with PyTorch[1] [32]. The coefficients in loss functions are set as $\lambda_\alpha = 1$, $\lambda_{FB} = 0.25$, $\gamma = 5 \times 10^4$, and $j = 4$. We use the Kaiming initializer [20] to initialize the network parameters. To prevent overfitting, we adopt the ResNet-50 [21] with group normalization [43] and weight standardization [33] pre-trained on ImageNet [10] as the backbone for the encoders of the propagating module and matting module. Meanwhile, we pre-train the propagating module on upsampled Adobe Image Matting [44] and Dinstinctions-646 datasets [34].

---

[1]The source code is available at https://github.com/QLYoo/LFPNet.

Table 2: The quantitative results on the Adobe Image Matting dataset. Note that "Whole" methods take the whole images for image matting and "Patched" methods take image patches for image matting. † denotes that the test-time augmentation is used during inference. The best results are highlighted in bold.

| | Methods | SAD | MSE | Grad | Conn |
| --- | --- | --- | --- | --- | --- |
| Whole | KNN-Matting [5] | 175.4 | 103.0 | 124.1 | 176.4 |
| | Closed-Form [24] | 168.1 | 91.0 | 126.9 | 167.9 |
| | AlphaGAN [30] | 52.4 | 30.0 | 38.0 | 53.0 |
| | DIM [44] | 50.4 | 14.0 | 31.0 | 50.8 |
| | IndexNet [28] | 45.8 | 13.0 | 25.9 | 43.7 |
| | AdaMatting [3] | 41.7 | 10.0 | 16.8 | — |
| | ContextNet [22] | 35.8 | 8.2 | 17.3 | 33.2 |
| | GCAMatting [26] | 35.3 | 9.1 | 16.9 | 32.5 |
| | FBAMatting† [13] | 25.8 | 5.2 | 10.6 | 20.8 |
| Patched | HDMatt [45] | 33.5 | 7.3 | 14.5 | 29.9 |
| | LFPNet (Ours) | 23.6 | 4.1 | 8.4 | 18.5 |
| | LFPNet† (Ours) | **22.4** | **3.6** | **7.6** | **17.1** |

We preprocess the images with several common data augmentation methods, including random affine transformation, random saturation transformation, random gray-scale transformation, random gamma transformation, random contrast transformation, and random composition. The images are randomly cropped to patches of dimensions 768×768, 640×640, 512×512, 448×448, 320×320. The trimap is generated from the alpha matte ground truth by random erosion and dilation with kernel sizes of 3 to 35 pixels. In addition, we randomly change the foreground regions in the trimap to the unknown regions.

We train the network with a batch size of 1 on the Adobe Image Matting dataset [44] using an NVIDIA GTX 1080Ti GPU. We use the RAdam optimizer [27] with a weight decay of $10^{-5}$ and betas of $(0.5, 0.999)$. The learning rate is fixed to $10^{-5}$ in the training procedure. The training procedure, which includes three stages. First, we train the matting module for 35 epochs. Then, we reinitialize the optimizer and train the decoders of the propagating module and the matting module for 10 epochs. Finally, we reinitialize the optimizer and fine-tune the whole network for 5 epochs.

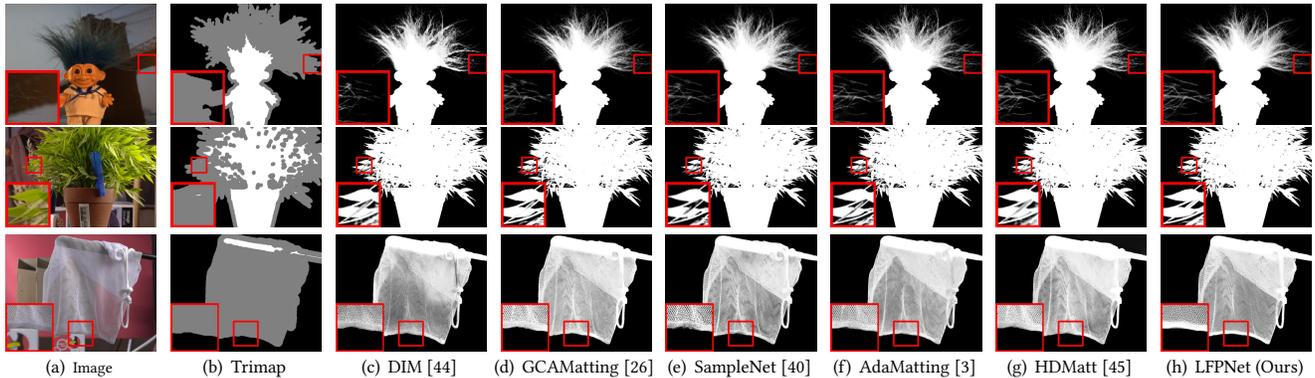

Figure 5: Qualitative comparison of the alpha matte results on the AlphaMatting dataset.

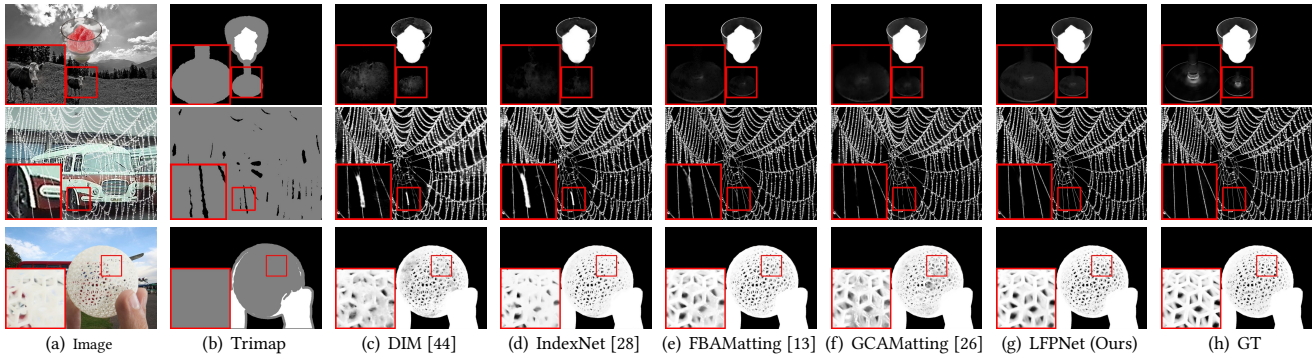

Figure 6: Qualitative comparison of the alpha matte results on the Adobe Image Matting dataset. GT stands for "Ground Truth".

### 5.3 Comparison to the State-of-the-art Methods

We compare LFPNet to other state-of-the-art methods on the AlphaMatting and Adobe Image Matting datasets. As shown in Tables 1 and 2, the proposed LFPNet achieves the best performance in terms of SAD, MSE, Gradient on the AlphaMatting dataset. LFPNet also achieves the best performance in terms of all metrics on the Adobe Image Matting dataset. Specifically, we use the patch-based crop-and-stitch inference [45], where the images are cropped into $1024 \times 1024$ patches before being fed to the network. Figures 6 and 5 give the qualitative results of the estimated alpha mattes, which indicates that the proposed LFPNet generates sharper alpha mattes and is more robust in local background interference.

### 5.4 Results on Real-world Images

LFPNet not only performs favorably against the state-of-the-art methods on the AlphaMatting and Adobe ImageMatting datasets, but also performs well on real-world high-resolution images. To demonstrate the performance of the proposed LFPnet, we collect some real-world high-resolution images from the Internet and annotate the corresponding trimaps. We use IndexNet [28], FBAMatting [13], GCAMatting [26] and LFPNet to estimate the alpha mattes. Since IndexNet, FBAMatting and GCAMatting process the whole picture, we implement these methods on the CPUs to avoid insufficient memory errors. Figure 7 gives the qualitative results of the estimated alpha mattes, LFPNet extracts finer image details while avoiding background interference.

### 5.5 Ablation Study

To compare the performance with different parameters of the propagating module, the sizes of the inner center image patch and the context image patch, we conduct the ablation study on the Adobe Image Matting dataset [44].

**Propagating Module.** To propagate long-range features for alpha matte estimation, the proposed LFPNet introduces the propagating module with Center-Surround Pyramid Pooling (CSPP). To evaluate the effectiveness of the propagating module, we compare the performance without the propagating module and with different architectures for extracting long-range features in the propagating module, including Non-Local Neural Network [42] and Atrous Spatial Pyramid Pooling (ASPP) [4]. As shown in Table 3, the propagating module with CSPP archives the best results in terms of SAD, MSE, Grad, and Conn.

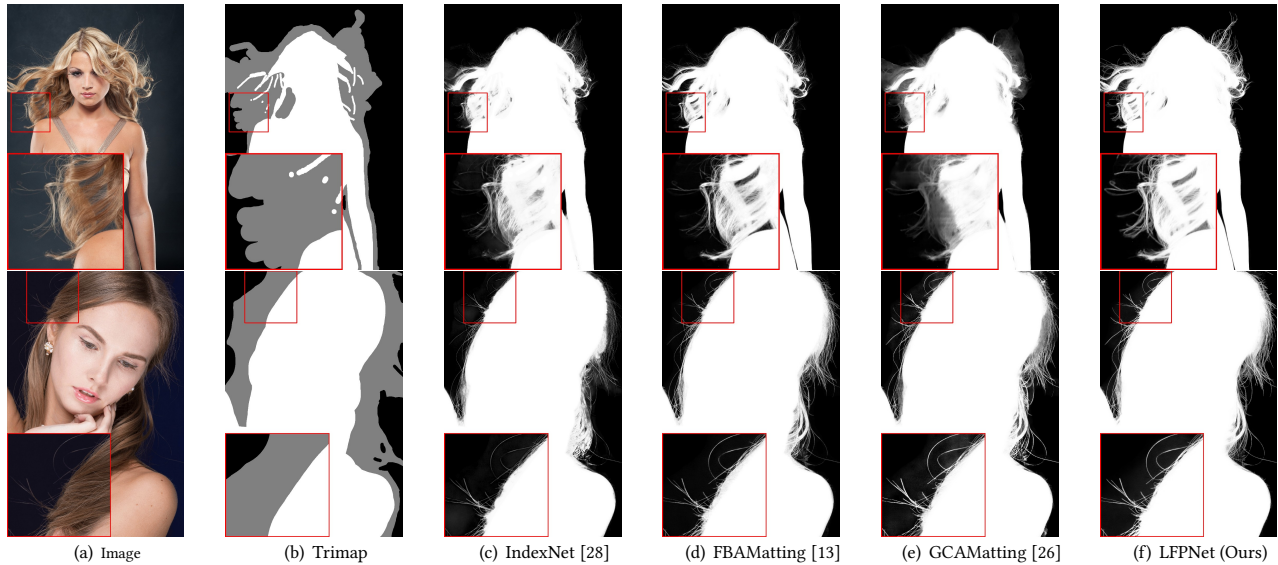

Figure 7: Qualitative comparison of the alpha matte results on real-world high resolution images.

Table 3: The effectiveness of the CSPP-based propagating module compared to the on the Adobe Image Matting dataset. Note that "P.M." represents "Propagating Module". The best results are highlighted in bold.

| Method | SAD | MSE | Grad | Conn |
|---|---|---|---|---|
| w/o P.M. | 30.1 | 5.4 | 11.0 | 26.6 |
| P.M. with Non-Local | 26.8 | 4.6 | 9.5 | 22.5 |
| P.M. with ASPP | 24.1 | 4.1 | 8.7 | 19.3 |
| P.M. with CSPP | **23.2** | **3.7** | **8.0** | **18.1** |

Table 4: The SAD, MSE, Grad, and Conn with different sizes of the inner center image patch on the Adobe Image Matting dataset. The best results are highlighted in bold.

| Inner Image Patch Size | SAD | MSE | Grad | Conn |
|---|---|---|---|---|
| 512 | 23.2 | 3.7 | 8.0 | 18.1 |
| 768 | 22.7 | 3.6 | 7.8 | 17.5 |
| 1024 | **22.4** | **3.6** | **7.6** | **17.1** |

**The Size of Inner Center Image Patch.** The proposed LFPNet adopts use the patch-based inference strategy to process high-resolution images. To investigate the effect of the size of the inner center image patch, we compare the performance with different sizes of the inner center image patch. The experimental results presented in Table 4 shows that larger inner center image patch size leads to accurate results, which indicates that rich image information benefits the alpha matte estimation.

**The Size of Context Image Patch.** To extract context features for alpha matte estimation, the propagating module of LFPNet takes the context image patch as input. To evaluate the impact of the size of the context image patch, we compare the performance with different sizes of the context image patch when fixing the size of the inner center image to 512 × 512. As shown in Table 5, the performance is better with large size of the context image patch, which indicates that the large context image patch provides more information that benefits the alpha matte estimation.

Table 5: The SAD, MSE, Grad, and Conn with different sizes of the context image patch on the Adobe Image Matting dataset. The best results are highlighted in bold.

| Context Patch Size | SAD | MSE | Grad | Conn |
|---|---|---|---|---|
| 512 | 24.1 | 3.9 | 8.5 | 19.0 |
| 768 | 23.5 | 3.7 | 8.2 | 18.4 |
| 1024 | **23.2** | **3.7** | **8.0** | **18.1** |

## 6 CONCLUSION

In this paper, we present Long-Range Feature Propagating Network (LFPNet) for natural image matting. Compared to existing deep learning methods that are limited by the small effective reception fields of convolutional neural networks, LFPNet learns the long-range context features outside the reception fields, which benefits the alpha matte estimation for the pixels in unknown regions by leveraging more pixels in the known foreground and background regions. Moreover, with the proposed center-surround pyramid pooling (CSPP), the smoothed long-range features can be explicitly propagated from the surrounding context image patch to the inner center image patch. Experiment results on the AlphaMatting and Adobe Image Matting datasets demonstrate that the proposed LFPNet outperforms the state-of-the-art methods.

## ACKNOWLEDGMENTS
This work was supported by the National Natural Science Foundation of China (Nos. 61872112, 62072141 and 61972167).